\documentclass[sigconf]{acmart}

\usepackage{siunitx}

\AtBeginDocument{%
  }

\setcopyright{acmlicensed}
\copyrightyear{2024}
\acmYear{2024}
\acmDOI{XXXXXXX.XXXXXXX}

\acmConference[Conference BIOKDD '24]{23rd International Workshop on Data Mining in Bioinformatics}{August 25,
  2024}{Barcelona, Spain}
\acmISBN{978-1-4503-XXXX-X/18/06}

\begin{document}

\title{gFlora: a topology-aware method to discover functional co-response groups in soil microbial communities}

\author{Nan Chen}
\orcid{0000-0003-4257-7265}
\affiliation{%
  \institution{University of Twente}
  \city{Enschede}
  \country{The Netherlands}
}
\email{n.chen@utwente.nl}

\author{Merlijn Schram}
\orcid{0009-0005-7396-9896}
\affiliation{%
  \institution{Netherlands Institute of Ecology (NIOO-KNAW)}
  \city{Wageningen}
  \country{The Netherlands}
}
\email{m.schram@nioo.knaw.nl}

\author{Doina Bucur} % I moved myself to the end
\orcid{0000-0002-4830-7162}
\affiliation{%
  \institution{University of Twente}
  \city{Enschede}
  \country{The Netherlands}}
\email{d.bucur@utwente.nl}

\begin{abstract}
Microorganisms such as bacteria perform critical functions in the soil ecosystem: they mediate essential carbon and nitrogen cycling processes in soils. To manage the health and functions of soils, it is important to understand which soil functions are controlled by which microbial taxa---but this taxon-to-function link is difficult to discover because of the size and complexity of the soil ecosystem. A feasible solution is to discover functional links at the level of group instead of individual, using observational data of both taxa abundance and soil function. We thus aim to learn the \emph{functional co-response group}: a group of taxa whose \emph{co-response effect} (the representative characteristic of the group showing the total topological abundance of taxa) co-responds (associates well statistically) to a functional variable. Different from the state-of-the-art method, we model the soil microbial community as an ecological co-occurrence network with the taxa as nodes (weighted by their abundance) and their relationships (a combination from both spatial and functional ecological aspects) as edges (weighted by the strength of the relationships). Then, we design a method called \emph{gFlora} which notably uses \emph{graph convolution} over this co-occurrence network to get the co-response effect of the group, such that the network topology is also considered in the discovery process. We evaluate gFlora on two real-world soil microbiome datasets (bacteria and nematodes) and compare it with the state-of-the-art method. gFlora outperforms this on all evaluation metrics, and discovers new functional evidence for taxa which were so far under-studied. We show that the graph convolution step is crucial to taxa with relatively low abundance (thus removing the bias towards taxa with higher abundance), and the discovered bacteria of different genera are distributed in the co-occurrence network but still tightly connected among themselves, demonstrating that topologically they fill different but collaborative functional roles in the ecological community.
\end{abstract}





\received{27 May 2024}
\received[revised]{XX XXX XXXX}
\received[accepted]{XX XXX XXXX}

\maketitle

%%%%%%%%%%%%%%%%%%%%%%%%%%%%%%%%%%%%%%%%%%%%%%%%%%%%%%%%%%%%%%%%%%%%%%%%%%%%%%%%%%%%%%%%%%%%%%%

\section{Introduction}

\begin{figure*}[tb] % This figure is out of place, but this is necessary to make it compile on the right page
  \centering
  \includegraphics[width=\linewidth]{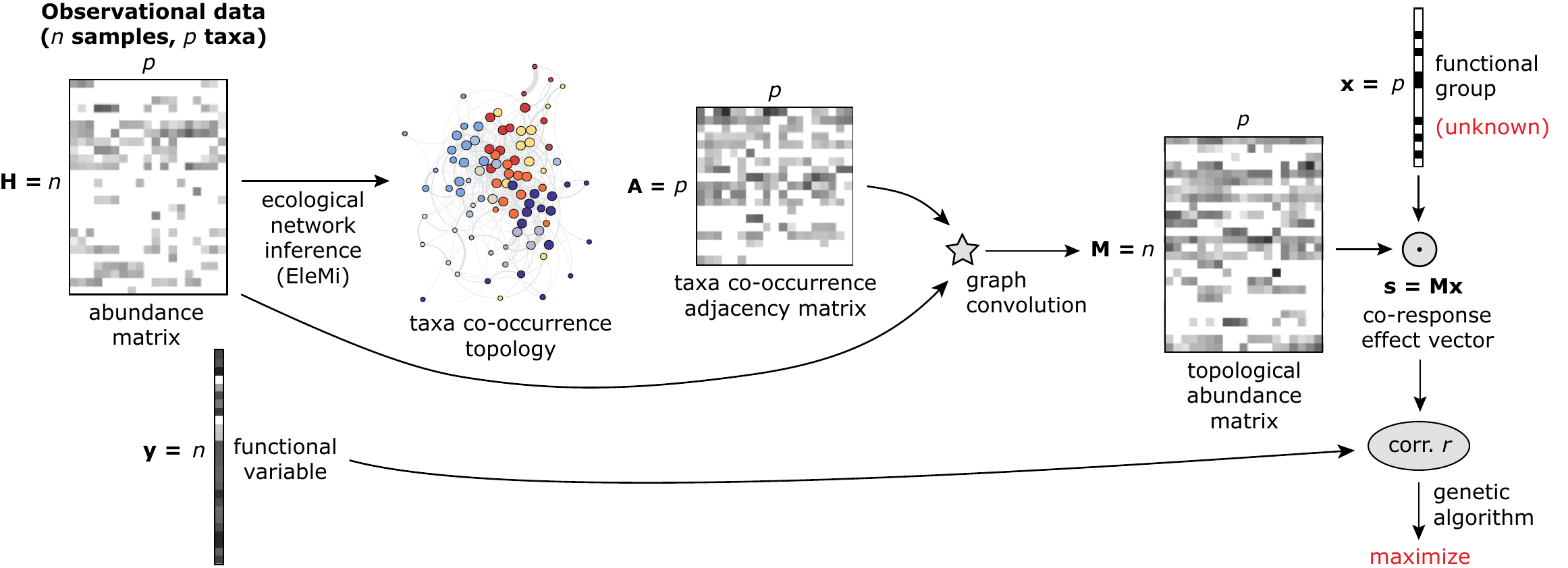}
  \caption{Overview of the gFlora method: from observational data (left) to optimizing the functional group $\mathbf{x}$ (right).}
  \label{fig:method}
\end{figure*}

The health of soils is linked to their biological condition: soil biology determines its capacity to function as a living system and to provide ecosystem services~\cite{EUlawsoilmon2023}. While essential to understand, predict, and control soil health and functions, a functional model for the soil microbial community is hard to discover, because a soil community is (1) large, with thousands of different bacterial and other microbial taxa, and (2) complex, with numerous unknown interactions between taxa which cannot be individually observed in situ, nor have been cultured in the lab beyond a few well-known taxa. The systematic discovery of the functions of individual taxa is particularly hard~\cite{fierer2017embracing}. The most widely used method to achieve this is through functional gene annotations and taxonomy, but these annotations are of low quality because of the size, complexity, and insufficient phenotypic characterization of the community~\cite{lobb2020assessment,schnoes2009annotation}. Furthermore, even if a genetic pathway is correctly discovered, this is not a guarantee that it is expressed by the organism in its living context.

Instead, a feasible solution to the complex problem of microbial functional characterization is to discover not individual taxa, but \emph{functional co-response groups}, and to do so by learning from \emph{observational data} which measures both the state of the soil, and the microbial community. A functional co-response group is defined as a group of taxa that co-respond to functional variables and whose \emph{co-response effect} (the representative characteristic of the group showing the total topological abundance of taxa) remains stable and associates well statistically with the \emph{functional variable} which reflects the metabolic result in the observational data~\cite{nelson2016global,shan2023annotation}. These functional co-response groups provide not only a simpler and more interpretable functional model of the soil biology, but are also computationally more feasible, since they reduce the complexity of the problem.

Ensemble Quotient Optimization (EQO)~\cite{shan2023annotation}, which is the state-of-the-art and only method to do so, showed that functional co-response group discovery from observational data is sound and feasible. However, EQO only uses \emph{group abundance} as the co-response effect: a group, whose total abundance is stable and correlates better with the functional variable, is assumed more likely to be a functional group; and they assume that the discovered taxa would be independent in abundance. This completely overlooks (1) \emph{the spatial aspect}: spatially closer taxa are easier and more efficient to engage in material exchange to support soil functions; (2) \emph{the functional ecological interactions} (such as competition or cooperation): group dynamic stability is remained by taxa replacing each other through these interactions. Both the spatial and ecological relationships can be learned from the observational data, and are modeled by an ecological co-occurrence network. For instance, spatially closer taxa tend to exhibit higher connectivity within the network, reflecting their shared environmental conditions. Furthermore, stronger ecological interactions also result in higher connectivity compared to weaker or no interactions. This dual encoding in the ecological co-occurrence network allows for a comprehensive understanding of both the spatial distribution and ecological dynamics within the soil ecosystem. We hypothesize that (1) taking this network (especially the network topology) into account would be helpful in functional co-response group discovery and (2) those discovered taxa might be distributed in the co-occurrence network but are still correlated instead of totally independent in abundance.

We thus propose the \emph{graph convolution-based functional co-response group discovery (gFlora)} method: a new, integrated, data-driven method to discover microbial functional co-response groups, which learns from (1) microbial abundances, (2) the functional variable, and also (3) the topology of the ecological co-occurrence network of microbes. Compared to EQO, gFlora also considers both the spatial and ecological relationships of taxa by computing a new co-response effect using graph convolution. Graph convolution propagates the abundance over the co-occurrence links, thus adding a "dynamic" view upon the soil ecosystem.

We evaluate gFlora on two real-world datasets describing bacteria and nematodes, and also measuring an important functional variable: potential nitrogen mineralization (PMN) which is crucial for nutrient cycling in the soil.

We summarize our results as follows:
\begin{itemize}
    \item gFlora performs significantly better than the EQO method on the testing datasets, on all performance metrics. Our new co-response effect of the discovered bacteria shows good correlation with PMN ($r=0.50$, $p<0.01$). We also offer a new option for controlling the size of the functional group, which provides better performance and quicker convergence of the optimization.
    \item We provide new evidence for taxa that are not extensively studied in the context of nitrogen cycling (N-cycling), thus enabling specialists to set up focused experiments to validate these taxa's functional behavior.
    \item We further show that our hypothesis holds: the co-occurrence network is useful for the discovery of functional co-response groups. The functionally important bacteria are distributed in the co-occurrence network (often in different clusters) and play different roles in the network. We thus see evidence that functional taxa would collaborate across clusters, likely on different aspects of the specific function.
\end{itemize}

Overall, gFlora could help ecologists better understand and manage soil functions. With gFlora, new hypotheses might be made and further validated, thus providing insights to intervene and regulate for healthier soil. The source code of gFlora is publicly accessible on GitHub: \url{https://github.com/nan-v-chen/gFlora}.

%%%%%%%%%%%%%%%%%%%%%%%%%%%%%%%%%%%%%%%%%%%%%%%%%%%%%%%%%%%%%%%%%%%%%%%%%%%%%%%%%%%%%%%%%%%%%%%

\section{Methods}

We provide an overview of our method in Fig. \ref{fig:method}; the steps in the method are described below. gFlora aims to learn a functional co-response group from observational data of $n$ soil samples, each measuring $p$ microbial taxa (the composition of the biological community), plus an indicator of the soil's ability to perform an important function. This input data (shown on the left in Fig. \ref{fig:method}) thus represents:
\begin{itemize}
    \item the abundance of individual taxa, in the \emph{abundance matrix} denoted $\mathbf{H}\in\mathbb{R}^{n\times{p}}$, and
    \item the function indicator per sample, in the \emph{functional variable} denoted $\mathbf{y}\in\mathbb{R}^{n}$.
\end{itemize}

The result of the learning process is the \emph{functional co-response group} $\mathbf{x}\in\mathbb{B}^{p}$: a Boolean vector for which 1 or 0 represent the presence or absence of a taxon in the functional group.

\subsection{Graph Convolution}

The microbial community (represented in raw form in the abundance matrix $\mathbf{H}$) can be modeled as a \emph{co-occurrence network}, in which the nodes are taxa and the edges are co-occurrence relationships. These co-occurrence relationships indicate a combination of (a) spatial relationships of a non-functional nature and (b) functional ecological interactions (such as competition or cooperation~\cite{connor2017using, yuan2021characteristics}). We know that the co-response effect of functional co-response groups can remain stable even though individual taxa may fluctuate in abundance, by replacing each other in space and time with the help of those ecological interactions~\cite{shan2023annotation}. Thus, taking these co-occurrence relationships (the network topology) into account can better capture the dynamic, adaptive stability of functional groups.

To achieve this, we compute the adjacency matrix $\mathbf{A}\in\mathbb{R}^{p\times{p}}$ of the co-occurrence network inferred using the most recent method EleMi~\cite{chen2024elemi}. EleMi infers this network from abundance data using multi-regression. Compared with earlier methods like SPIEC-EASI, it was shown to be more robust and provide a clearer cluster structure. In the co-occurrence network, the nodes are taxa and the undirected weighted edges are co-occurrence relationships (there are no self-loops).

Using the adjacency matrix $\mathbf{A}$ of the co-occurrence network, the original abundance matrix $\mathbf{H}$ is then converted into a \emph{topological abundance matrix} $\mathbf{M}\in\mathbb{R}^{n \times p}$ using graph convolution \cite{zhang2019graph}. Graph convolution is an operation that captures dynamic relationships and the topological information of the network, by aggregating information from a taxon's neighbors. The expression for $\mathbf{M}$ is:

\begin{equation}\label{eq:M}
    \mathbf{M} = \mathbf{H} \left( \widetilde{\mathbf{D}}^{-\frac{1}{2}} \widetilde{\mathbf{A}} \widetilde{\mathbf{D}}^{-\frac{1}{2}} \right),
\end{equation}
where $\widetilde{\mathbf{A}}=\mathbf{A}+\mathbf{I}$ is the adjacency matrix $\mathbf{A}$ plus the identity matrix $\mathbf{I}$ (this addition ensures that, along with aggregating the neighbor information, the node's own features are also preserved). $\widetilde{\mathbf{D}}$ is the diagonal weighted degree matrix of $\widetilde{\mathbf{A}}$.

\subsection{Functional Co-Response Group Discovery}

Similar to EQO \cite{shan2023annotation}, given the topological abundance matrix $\mathbf{M}$, gFlora computes the \emph{co-response effect vector} $\mathbf{s}\in\mathbb{R}^{n}$ with:
\begin{equation}
    \mathbf{s} = \mathbf{M}\mathbf{x}.
\end{equation}
The $i$th element of $\mathbf{s}$ is the group co-response effect in the $i$th sample. $\mathbf{x}$ models the functional co-response group (which is unknown, and the learning goal): a Boolean vector of size $p$ (the number of taxa). For each taxon in $\mathbf{x}$, a value of 1 represents presence, and 0 represents absence in the group. Intuitively, $\mathbf{s}$ is a vector of size $n$ (the number of samples) showing the total topological abundance of taxa in a given group $\mathbf{x}$, per sample. With a proposed $\mathbf{x}$, the co-response effect vector $\mathbf{s}$ is computed out of this $\mathbf{x}$ and $\mathbf{M}$, which was calculated out of the taxa abundance by Eq.~(\ref{eq:M}).

We then need to examine the correlation between the co-response effect vector $\mathbf{s}$ and the functional variable $\mathbf{y}$. Therefore, the objective function is to maximize the Pearson's correlation coefficient $r$ between $\mathbf{s}$ and $\mathbf{y}$:
\begin{equation}
    \max r.
\end{equation}
This is actually a function of x. To simplify the algebra, $\mathbf{s}$, $\mathbf{y}$ and the columns of $\mathbf{M}$ can be scaled to $\mathbf{s_0}$, $\mathbf{y_0}$ and $\mathbf{M_0}$ by subtracting their means without affecting correlations. The objective function becomes:
\begin{align}\label{eq:obj1}
    \max \frac{Cov(\mathbf{s_0},\mathbf{y_0})}{\sigma(\mathbf{s_0})\sigma(\mathbf{y_0})} \nonumber
    &= \max \frac{\mathbf{s_0}^T\mathbf{y_0}}{\sqrt{\mathbf{s_0}^T\mathbf{s_0} \cdot \mathbf{y_0}^T\mathbf{y_0}}} \nonumber \\
    &= \max \frac{\mathbf{x}^T\mathbf{M_0}^T\mathbf{y_0}}{\sqrt{\mathbf{x}^T\mathbf{M_0}^T\mathbf{M_0}\mathbf{x}}}.
\end{align}
 
The genetic algorithm \cite{scrucca2013ga} can be used to solve this combinatorial optimization problem. Specifically, 1 and 0 in $\mathbf{x}$ can be regarded as loci, and different combinations are different genotypes. Eq.~(\ref{eq:obj1}) is the fitness function.

There is a risk of overfitting because if the group size is bigger, it is easier to find a solution with a good fit. To mitigate the effect of overfitting, a penalty (regularization) should be put on the group size. In EQO, the best group size was decided separately using the Akaike Information Criterion (AIC), calculated as $-2k - \ln(L_k)$, where $L_k$ is the regression likelihood and $k$ is the group size. A smaller AIC indicates a better trade-off between goodness of fit and model complexity. Then, the fitness function is:
\begin{equation}\label{eq:obj2}
    \frac{\mathbf{x}^T\mathbf{M_0}^T\mathbf{y_0}}{\sqrt{\mathbf{x}^T\mathbf{M_0}^T\mathbf{M_0}\mathbf{x}}} - \alpha\max(\|\mathbf{x}\|_1-k_{opt}, 0),
\end{equation}
where $\alpha$ is a very big number, $\|\mathbf{.}\|_1$ is the $\ell_1$ norm, and $k_{opt}$ is the best group size. In this way, if the group size is bigger than $k_{opt}$, there will be a very big loss for the fitness function. 

gFlora employs the same criteria in deciding the group size. However, a predetermined group size $k$ would reduce the diversity of the population in the genetic algorithm, which may lead to local optima. For this, gFlora\_l1 additionally offers a new option of regularization to the fitness function:
\begin{equation}\label{eq:obj3}
    \frac{\mathbf{x}^T\mathbf{M_0}^T\mathbf{y_0}}{\sqrt{\mathbf{x}^T\mathbf{M_0}^T\mathbf{M_0}\mathbf{x}}} - \mu\|\mathbf{x}\|_1,
\end{equation}
where $\mu$ is a penalty parameter. With this fitness function, the group size is more flexible, which can enhance the diversity of the population and may result in a better solution.

\subsection{The functional group as a network}

\begin{figure}[htb]
  \centering
  \includegraphics[width=\linewidth]{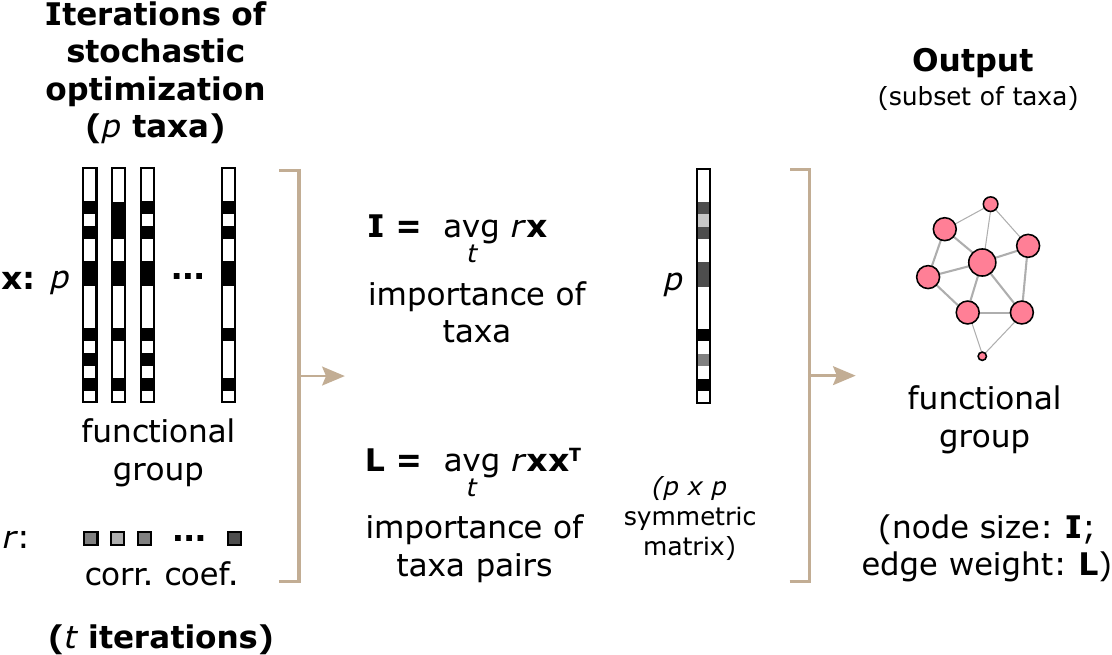}
  \caption{The gFlora output: a functional co-response group of taxa as an undirected network}
  \label{fig:method_output}
\end{figure}

\begin{figure*}[hbt] % figure out of place, but necessary to compile on the right page
  \centering
  \includegraphics[width=\textwidth]{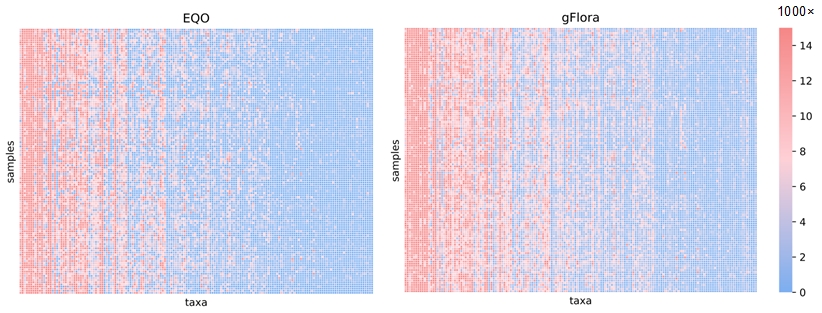}
  \caption{Comparison of $\mathbf{M}$ between EQO (the original relative abundance) and gFlora (the updated topological relative abundance). }
  \label{fig:comp_m}
\end{figure*}

Since the genetic algorithm is stochastic (probability-driven)~\cite{kromer2013randomness}, instead of giving an "optimal" $\mathbf{x}$, we repeat our algorithm several times and summarize the functional importance of all taxa as the vector $\mathbf{I}$ (as shown in Fig. \ref{fig:method_output}):
\begin{equation}
    \mathbf{I} = \underset{t}{\mathrm{avg}}\ r\mathbf{x}.
\end{equation}
Intuitively, $\mathbf{I}$ is the average of all solutions for $\mathbf{x}$, weighted by $r$, the correlation coefficient between the co-response effect of the functional group $\mathbf{x}$ and the functional variable. 

In order to investigate how much the taxa pairs contribute to the functional co-response group, their importance is computed as the adjacency matrix $\mathbf{L}$, such that the final output of gFlora is the functional group as a network:
\begin{equation}
    \mathbf{L} = \underset{t}{\mathrm{avg}}\ r\mathbf{x}\mathbf{x}^T.
\end{equation}

\subsection{Datasets and Experimental Settings}

\begin{figure}[htb]  % figure out of place, but necessary to compile on the right page
  \centering
  \includegraphics[width=7.7cm]{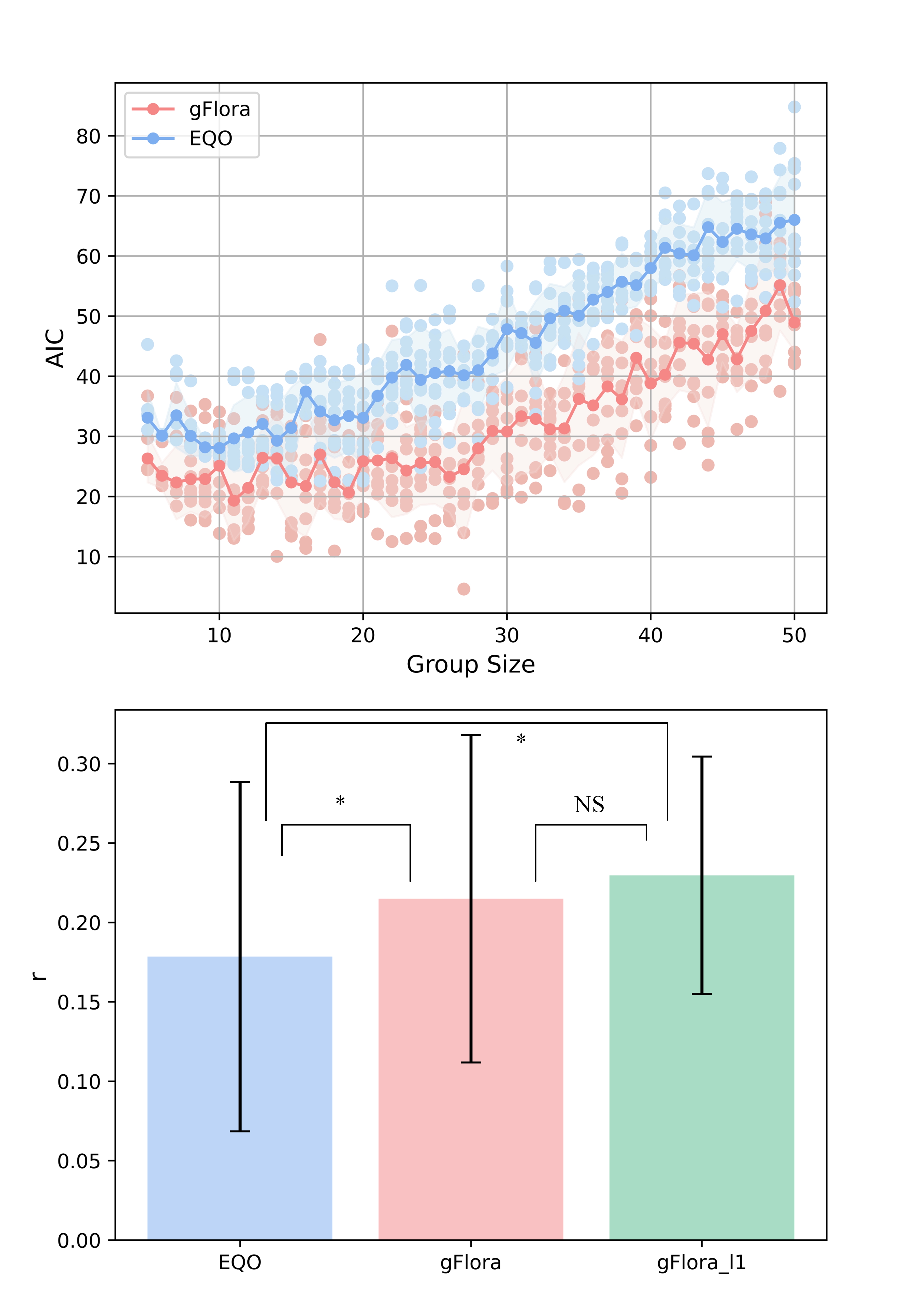}
  \caption{Results on bacteria. Group size selection with AIC (top, dots are AICs got in repeated experiments, line plots represent the change of mean values) and comparison of performance (bottom, bars are mean values and errors are standard deviations). *: significant; NS: not significant.}
  \label{fig:res_bac}
\end{figure}

\begin{figure}[htb]  % figure out of place, but necessary to compile on the right page
  \centering
  \includegraphics[width=7.7cm]{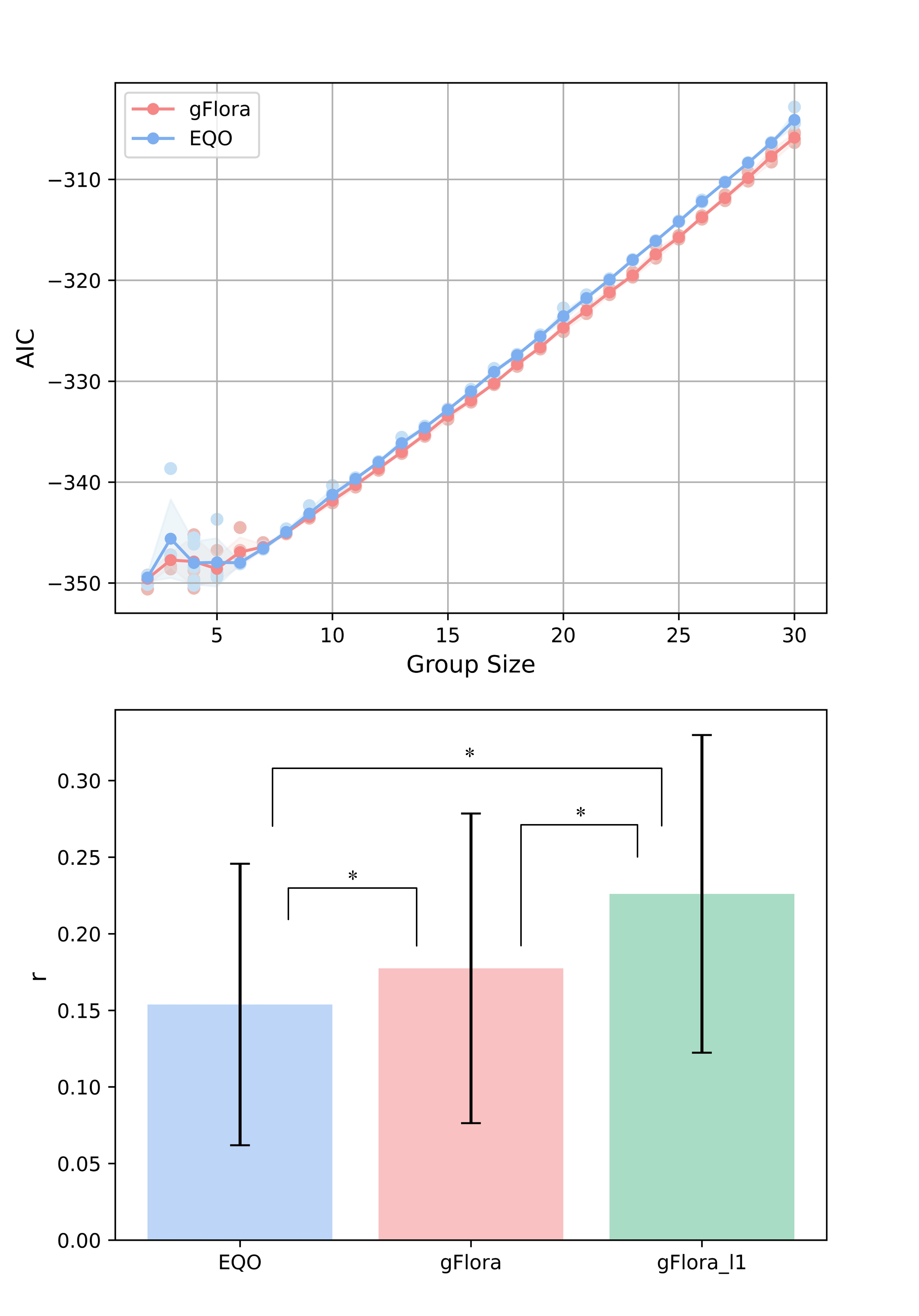}
  \caption{Results on nematodes. Group size selection with AIC (top, dots are AICs got in repeated experiments, line plots represent the change of mean values) and comparison of performance (bottom, bars are mean values and errors are standard deviations). *: significant.}
  \label{fig:res_nem}
\end{figure}

\subsubsection*{Microbiome datasets}
The soil samples were collected in the Netherlands before 2021. We experimented with microscopic organisms (microflora and microfauna) with body sizes on the opposite sides of the micro scale: \textbf{bacteria} are the smallest (body size \SI{1}{\micro\metre}) and \textbf{nematodes} are the largest (body size up to \SI{100}{\micro\metre}). The bacterial data is publicly available as a dataset~\cite{vanrijssel2022vitalsoilsdataset} first described in~\cite{vanrijssel2022vitalsoils}. We used the data from the soil type best represented in the dataset, i.e., marine clay. For this soil type, 46 arable fields were paired (in terms of organic vs.\ conventional management) and also matched in their local environmental factors. Per field, one soil sample was collected from three subsites, at 5–15 cm depth (the topsoil was excluded to avoid readings which vary with daily weather). DNA was extracted from bulk soil and the community composition of bacteria was determined by amplifying the V4-region of the 16S rRNA gene with polymerase chain reactions (PCRs), then sequenced (250 base pairs, paired end). Amplicon sequence variants (ASVs) with very low abundance were removed. Nematodes were extracted using gravity-sieving, followed by sucrose centrifugation-flotation, and then counted on an inverted microscope.

The optimization of gFlora is based on a genetic algorithm, thus managing the search space is crucial to ensure computational efficiency. We highly recommend reducing the number of taxa in the community to less than 500, otherwise the discovery of functional co-response groups may become too impractical due to the explosion of the search space size. The taxonomy of both bacteria and nematodes was originally assigned at the species level by consulting a database~\cite{vanrijssel2022vitalsoils}. We directly used nematode species as input to our algorithm, but we grouped bacterial taxonomic composition to the level of genera because the number of bacterial species is very large. To further constrain the search space and the computational complexity, we removed extremely ``sparse'' taxa (as suggested by EQO \cite{shan2023annotation}, those whose ratio of zero-abundance across soil samples is over 80\% do not carry a lot of useful information in the statistical sense). Nevertheless, a lot of relatively rare taxa are still available to be chosen. After this, the dimensions of the abundance matrices of bacteria and nematodes were $125\times164$ and $126\times92$, respectively. The abundance matrices are normalized with the Cumulative-Sum Scaling (CSS, ~\cite{paulson2013differential}) method (also as indicated in EQO~\cite{shan2023annotation}).

We chose the \textbf{potential nitrogen mineralization} (PMN) as our functional variable. This is the measured difference in ammonium (NH4+) and nitrate (NO3-) in 24-day incubation, per soil sample. The production of these bioavailable nitrogen is of great ecological importance: it enables nutrient cycling in both natural and managed soils~\cite{bardgett2005biology}. In fact, different processes (like nitrogen fixation and nitrification) in N-cycling also contributed to the change of NH4+ and NO3-, so that PMN is actually a combinational indicator of these processes.

In order to evaluate the effectiveness and generalization ability of our method, the data was divided into training and testing datasets (50\%-50\%, as in the related work~\cite{shan2023annotation}). The algorithm was trained on the training datasets to get the functional co-response groups. These groups were then used to compute the correlation $r$ between the group's co-response effect and the functional variable on the corresponding testing datasets. We did not use classic cross-fold validation because of the limited sample size (we need enough samples to train and test the model). Instead, we repeated this process 100 times to mitigate the influence of randomness on the evaluation. Due to the sensitivity of correlation to the distribution of the data, stratified sampling on the functional variable (separated into 10 quantiles, sampled proportionately) was used in the sampling process to ensure the representativeness of the training sample.

\subsubsection*{Experimental settings}
Our method contains the following software building blocks, and their configured parameters. 

EleMi~\cite{chen2024elemi} is the state-of-the-art in inferring soil ecological networks, which model the co-occurrence of the taxa. EleMi was implemented in Python (version 3.8.5), and is publicly available. It is configured here with the penalty parameter for the $\ell_1$ norm (of the Lasso term in the multi-regression) $\mu_1=0.1$ and likewise for the Frobenius norm (of the Ridge term) $\mu_2=0.01$; these values are chosen based on the hyperparameter tuning done on the same dataset in the original description of EleMi~\cite{chen2024elemi}.

The genetic algorithm was implemented using the R package GA (version 3.2.3)~\cite{scrucca2013ga} with the crossover probability set to 0.8, the mutation probability set to 0.1, and a population size of 200 (the same as in the related work~\cite{shan2023annotation}). The maximum number of iterations was set to 500. Early stop choices are also available using maximum fitness and the number of consecutive generations without any improvement in the best fitness value. Parallel computing is enabled by the R package doParallel (version 1.0.17). The penalty $\alpha$ in the original fitness function was set as the root of the maximum value of double-precision float in the current R environment ($.Machine\$double.xmax$). The $k_{opt}$ parameter (the best group size) was optimized in the range of 2 to 50 according to the AIC (The experiment is repeated 10 times for each group size). $\mu$ for $\ell_1$ norm is optimized in the cross-validation process within $\{1/30, 1/40, ..., 1/100\}$ on the training dataset.

To fully utilize all the data, when computing the importance of each taxon in the functional co-response group, we ran the algorithms on the entire dataset and repeated it 10 times. Besides, all statistical analyses were performed using paired two-sample t-test in R (version 4.3.1) with normality test. A significance threshold of 0.05 was applied to determine statistical significance.

%%%%%%%%%%%%%%%%%%%%%%%%%%%%%%%%%%%%%%%%%%%%%%%%%%%%%%%%%%%%%%%%%%%%%%%%%%%%%%%%%%%%%%%%%%%%%%%

\section{Results and Discussion}

\subsubsection*{The effect of graph convolution on $\mathbf{M}$}
A visual comparison of $\mathbf{M}$ between EQO and gFlora is shown in Fig. \ref{fig:comp_m}. The original sparse abundance matrix was updated through graph convolution with co-occurrence networks, strengthening the group effect by aggregating the information of the neighborhood. As can be observed in Fig. \ref{fig:comp_m}, zeroes are replaced by some small numbers, which makes it possible for some rare taxa to be also considered.

\subsection{Algorithm performance}

The results of comparison experiments are shown in Fig. \ref{fig:res_bac} and Fig. \ref{fig:res_nem}. AIC is used to choose the best group size in both EQO and gFlora. A smaller AIC indicates a better trade-off between goodness of fit and model complexity. When the group size is the same on the same dataset, a smaller AIC suggests a better predictive power. 

On the dataset of \textbf{bacteria} genera, our new method gFlora shows a smaller AIC, which indicates that the functional co-response group discovered by our method can better predict the functional variable PMN. EQO and gFlora obtain the smallest AIC with group size of 10 and 11 respectively (from Fig. \ref{fig:res_bac}, top). From Fig. \ref{fig:res_bac} (bottom), the correlation coefficient $r$ with the functional variable on the testing dataset of gFlora ($mean=0.21$, $std=0.10$) is significantly better than EQO ($mean=0.18$, $std=0.11$). gFlora with $\ell_1$ norm (gFlora\_l1, $mean=0.23$, $std=0.07$) also significantly outperforms EQO. No significant difference is found between the performance of gFlora and gFlora\_l1. This may be due to the small effect of group size on the performance of gFlora (when the group size is below 30), as shown by Fig. \ref{fig:res_bac} (top).

On the dataset of \textbf{nematode} species, the chosen group size of gFlora and EQO is both 5. Similarly, the performance of gFlora ($mean=0.18$, $std=0.10$) and gFlora\_l1 ($mean=0.23$, $std=0.09$) is significantly better than that of EQO ($mean=0.15$, $std=0.10$). Besides, gFlora\_l1 performs significantly better than gFlora because of a bigger effect of group size on performance. We note that the performance of functional co-response group discovery algorithms on bacteria is slightly better than on nematodes, which may be attributed to that bacteria are more directly linked to N-cycling processes~\cite{zheng2019soil}.

\subsubsection*{The effect of $\ell_1$ norm in gFlora\_l1}
\begin{figure}[htb]
  \centering
  \includegraphics[width=\linewidth]{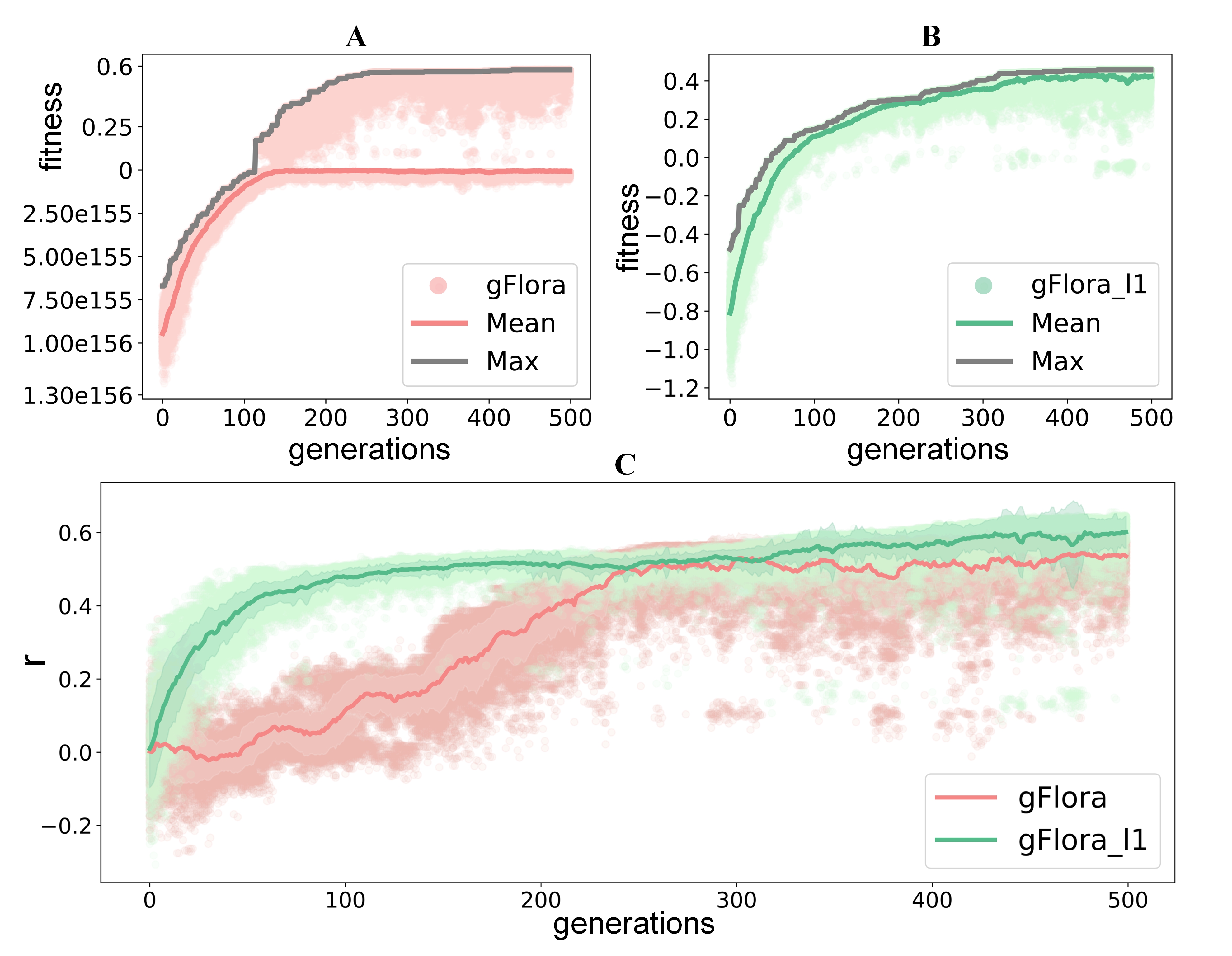}
  \caption{Generations of the genetic algorithm on bacteria. Dots are the population in each generation.}
  \label{fig:iter_bac}
\end{figure}

\begin{figure}[htb]
  \centering
  \includegraphics[width=\linewidth]{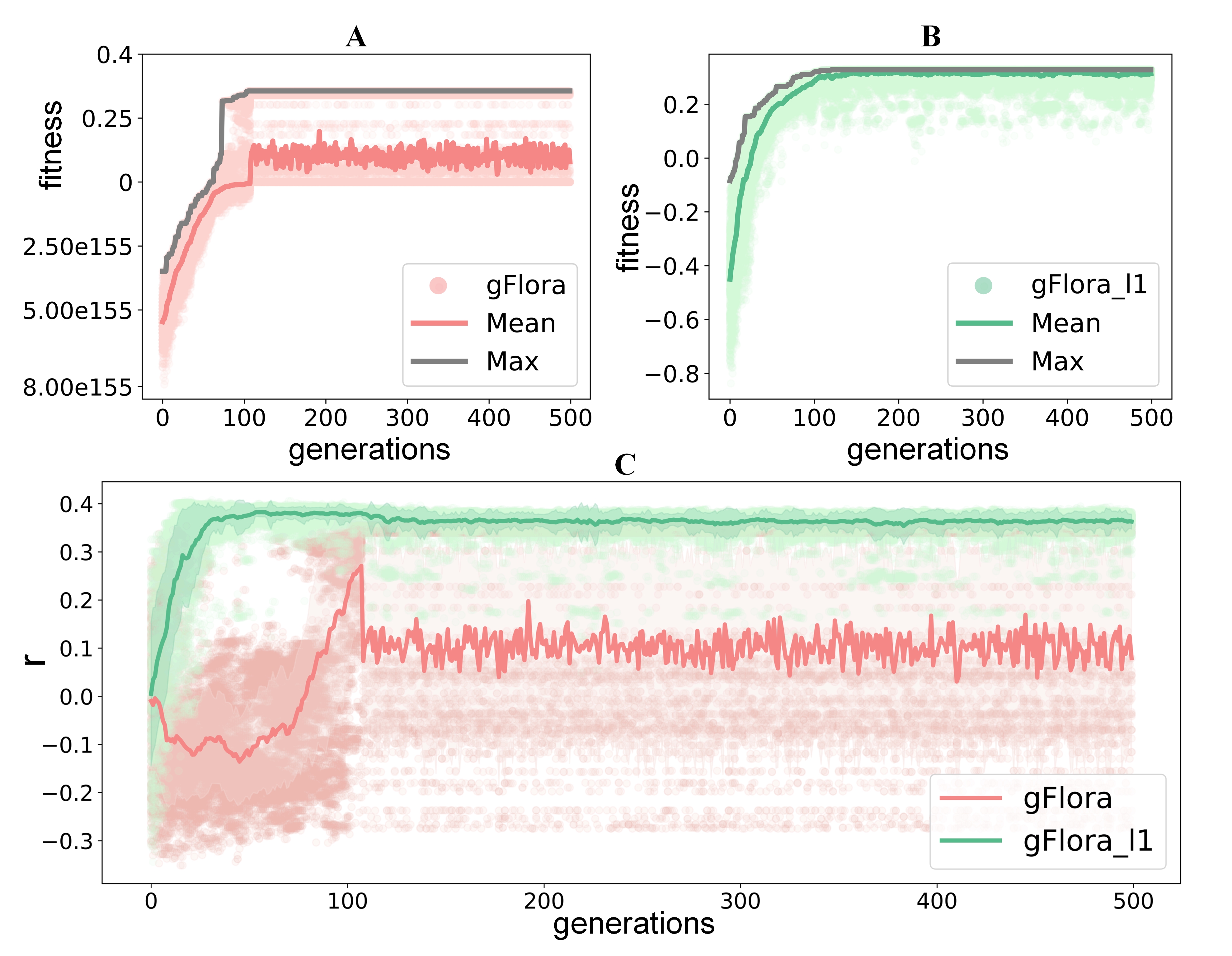}
  \caption{Generations of the genetic algorithm on nematodes. Dots are the population in each generation.}
  \label{fig:iter_nem}
\end{figure}

To further investigate the effect of the new option of regularization ($\ell_1$ norm) in the fitness function, we recorded the changes in fitness and $r$ in different generations, as is shown in Fig. \ref{fig:iter_bac} and Fig. \ref{fig:iter_nem}. It can be observed in the fitness curves (A and B in Fig. \ref{fig:iter_bac} and Fig. \ref{fig:iter_nem}) that there is a big gap between the Max curve (grey) and the Mean curve (red and green). This is because gFlora struggles to fit the predefined best group size ($k_{opt}$) more than getting a bigger fitness. Thus, the diversity of the population in group size of gFlora is much smaller than that of gFlora\_l1. As such, gFlora\_l1 exhibits faster convergence and gets better performance on $r$ compared to gFlora (Fig. \ref{fig:iter_bac}C and Fig. \ref{fig:iter_nem}C).

However, due to the parameter-tuning process needed in gFlora\_l1, its computational complexity is bigger than that of gFlora (for gFlora and EQO, 50 repetitions to decide $k_{opt}$ plus 100 iterations for comparison; for gFlora\_l1, 100 iterations for comparison with 7 repetitions to tune $\mu$ in each iteration). Besides, if the group size does not have a big effect on AIC, there would not be a significant increase in performance. Therefore, in practice, we recommend a pre-test with varying group sizes to decide whether to use gFlora\_l1 or not.

\subsection{Important taxa}

\begin{figure}[htb]
  \centering
  \includegraphics[width=\linewidth]{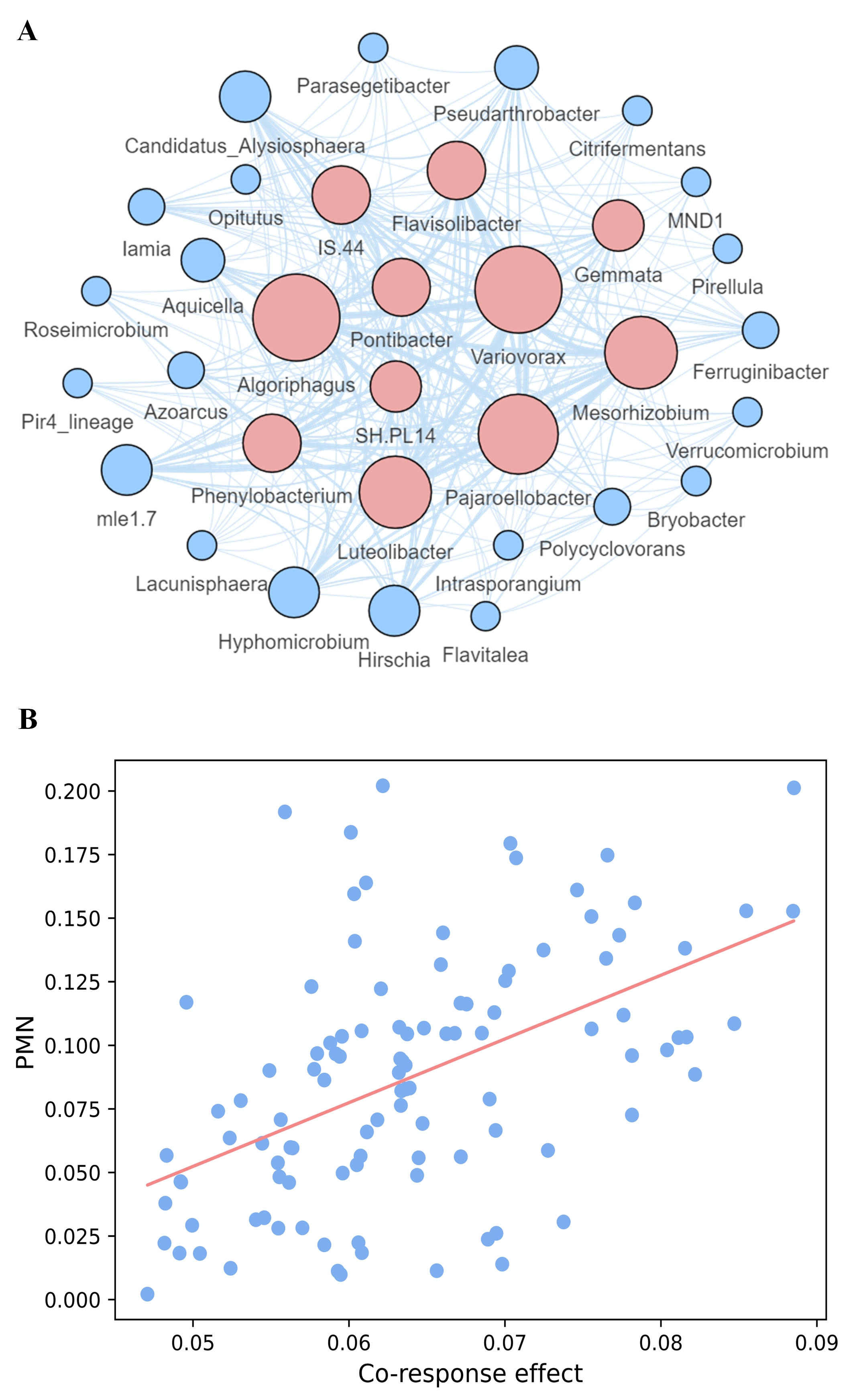}
  \caption{Functionally important bacteria. A) All bacteria with a functional role. The most important 11 bacteria are red; the rest are blue. The size of the nodes represents their importance and the width of edges is the importance of node pairs. B) The correlation between the co-response effect of the most important 11 bacteria and PMN across samples.}
  \label{fig:taxa}
\end{figure}

Nitrogen mineralization (also known as ammonification) in marine clay sediments is a crucial process involving the conversion of organic nitrogen into inorganic forms that plants can absorb, such as ammonium (NH4+) and nitrate (NO3-)~\cite{schimel2004nitrogen}. In our case, the functional variable PMN is measured using the difference in NH4+ and NO3- in 24-day incubation per soil sample. However, other processes, such as nitrogen fixation, also produce NH4+. Besides, nitrification can convert NH4+ to NO3-. Therefore, PMN is actually a combinational indicator of these processes, and all influences from these processes should be considered, for an accurate interpretation. These processes play a key role in N-cycling and the bioavailable nitrogen produced by these processes could support plant growth and soil fertility. Discovering key taxa related to the production of these bioavailable nitrogen is crucial for understanding soil health and ecosystem functioning. By identifying these microorganisms, we can develop targeted strategies to promote N-cycling, enhance soil fertility, and mitigate environmental impacts such as nitrogen pollution. Moreover, insights into these taxa can inform sustainable agricultural practices and biotechnological innovations aimed at optimizing nutrient management in soils.

All \textbf{bacteria} genera related to PMN discovered by gFlora are shown in Fig. \ref{fig:taxa}A as a network in which 11 bacteria with the highest importance in vector $\mathbf{I}$ are recognized as the most important and highlighted. We now discuss these findings, in light of what functional information is available in the literature (when such information exists), and the implications of these findings.

Among the bacterial genera in Fig. \ref{fig:taxa}A, \textit{Flavisolibacter}, \textit{Pontibacter} and \textit{Algoriphagus} belong to the phylum Bacteroidota (also known as Bacteriodetes), members of which have genes for nitrogen fixation and could therefore influence the production of bioavailable nitrogen~\cite{inoue2015distribution, madegwa2021land}. Species in the genera \textit{Mesorhizobium} and \textit{Phenylobacterium} are also involved in nitrogen fixation~\cite{li2024phylogenomic, jiao2023biochar, mehmood2022enrichment}. Both genera belong to the class Alphaproteobacteria, most members of which can always support this process~\cite{tsoy2016nitrogen}. \textit{Gemmata} and \textit{SH.PL14} belong to the phylum Planctomycetota, whose members are mostly capable of ammonium oxidation (part of nitrification), especially in marine soil~\cite{van2015anammox, wang2020response, glockner2003complete}. \textit{Variovax}, \textit{IS.44}, \textit{Pajaroellobacter} and \textit{Luteolibacter} are not as extensively studied in the context of N-cycling as other bacterial genera. Specifically, \textit{Pajaroellobacter} belongs to phylum Myxococcota, class Polyangia, members of which are usually predatory bacteria. So, how it influences the production of bioavailable nitrogen still needs to be explored. Note that \textit{Pajaroellobacter} and \textit{Flavisolibacter} were not recognized to be important by EQO. In particular, the relative abundance of \textit{Pajaroellobacter} is much smaller than other taxa, but nevertheless it appears to play an important role in the functional co-response group in gFlora, as shown in Fig. \ref{fig:taxa}A. Thus, the graph convolution does important work by updating the relative abundance with topological information of the co-occurrence network. Overall, this finding may raise the need to study \textit{Pajaroellobacter} in supporting N-cycling.

Given the complexity of microbial interactions, the contribution of \emph{groups of microorganisms} to a specific function might be different from their contribution as \emph{individuals}. However, the aforementioned studies on bacterial functioning have often focused on the individual level (primarily gene-related), rather than considering them as groups. Additionally, interventions targeting individual bacteria can have \emph{cascading effects} throughout the microbial community. Therefore, our method has the potential to discover some "untraditional" taxa related to a specific function so that could provide a new and broader perspective in the study of soil functions.

As expected, the co-response effect of these 11 bacteria is indeed significantly correlated (Fig. \ref{fig:taxa}B, $r = 0.50$, $p < 0.01$) with the functional variable PMN. As such, implementing appropriate interventions on these important bacteria can potentially enhance N-cycling.

\begin{figure}[htb]
  \centering
  \includegraphics[width=\linewidth]{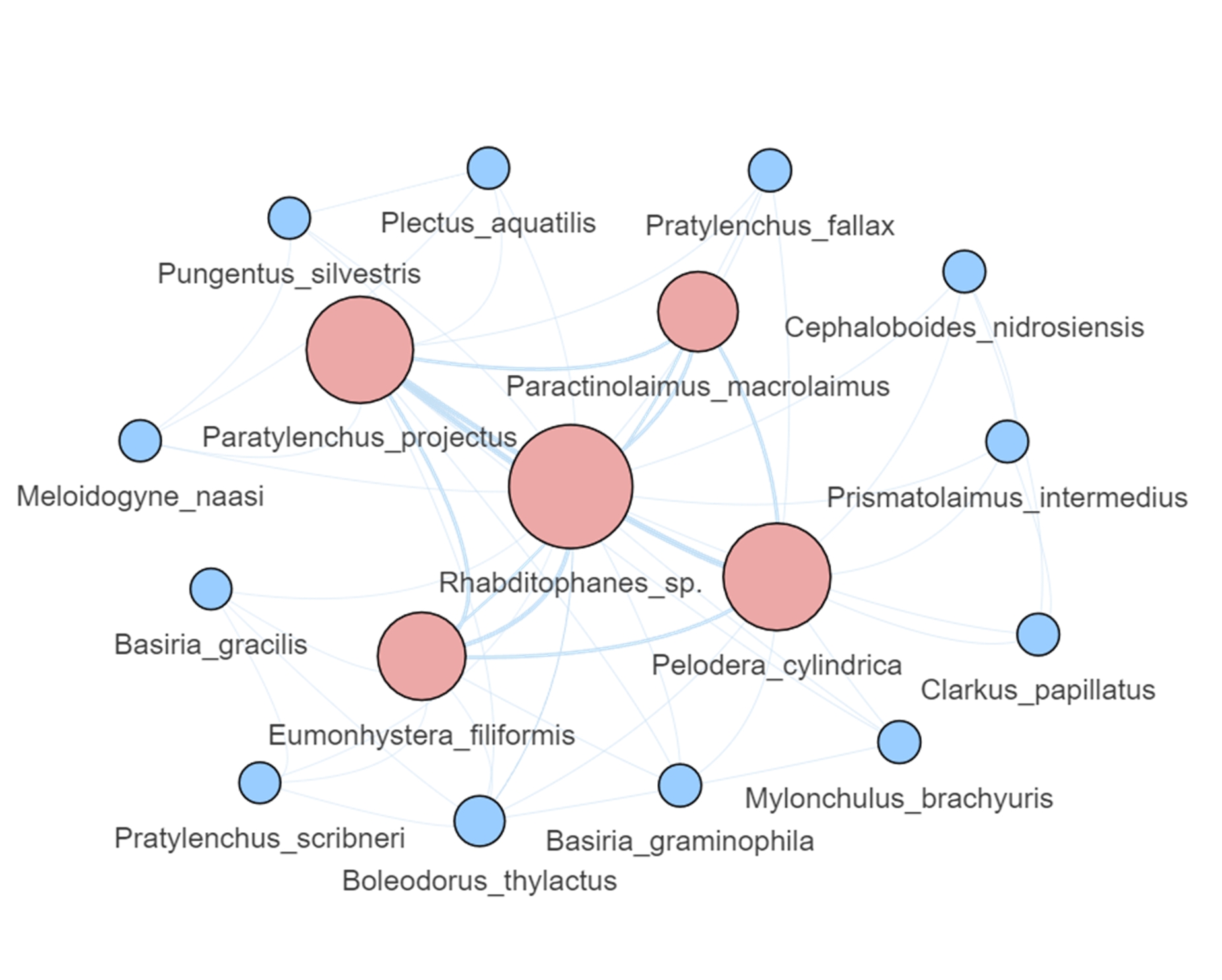}
  \caption{Functionally important nematodes. The most important 5 nematodes are red; the rest are blue. The size of the nodes represents their importance and the width of edges is the importance of node pairs.}
  \label{fig:taxaN}
\end{figure}

\begin{figure*}[htb]
  \centering
  \includegraphics[width=\textwidth]{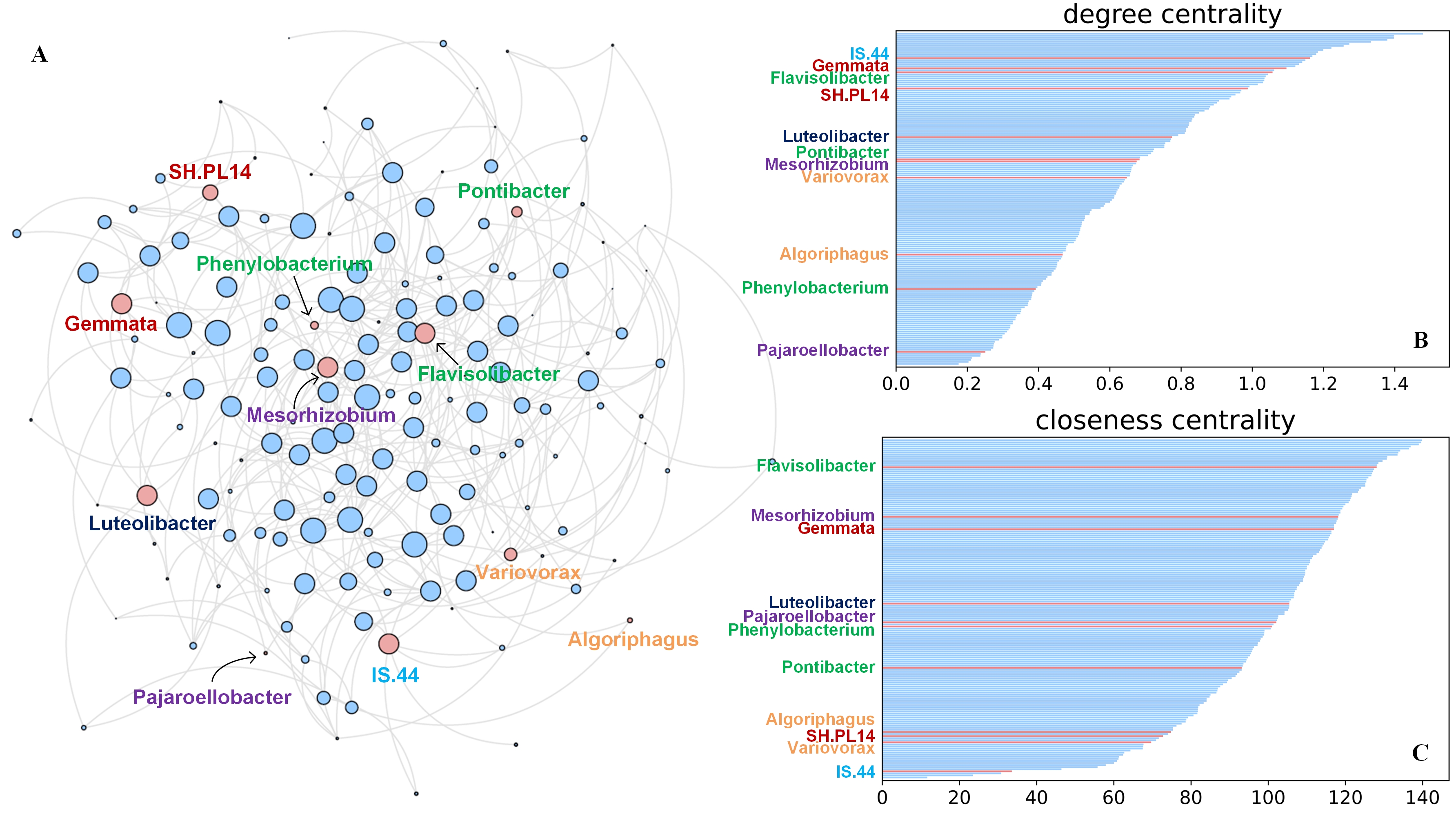}
  \caption{A) The most important 11 bacteria in the co-occurrence network (only edges with weight over 0.05 are shown). The most important 11 bacteria are colored in red while the rest are in blue (different colors of labels represent different clusters detected). The size of nodes denotes their mean relative abundance. B) The degree centrality (weighted) and C) The closeness centrality (weighted) of all nodes (bacteria).}
  \label{fig:taxa_in_co-occur}
\end{figure*}

Fig. \ref{fig:taxaN} shows all \textbf{nematode} species discovered by gFlora as a network. \textit{Rhabditophanes\_sp}, \textit{Pelodera\_cylindric} and \textit{Eumonhystera\_filiformis} are all bacterivores (feed on bacteria)~\cite{mahboob2023comparative, dulovic2020rhabditophanes, yeates1993feeding}, which has been proven to enhance the nitrogen mineralization of bacteria~\cite{ingham1985interactions}. Note that \textit{Eumonhystera\_filiformis}, whose relative abundance is small, was not recognized to be important by EQO, so the study of rare nematodes can also benefit from the application of graph convolution, enhancing our understanding of microbial and invertebrate interactions and ecological dynamics. \textit{Paractinolaimus\_macrolaimus} is a predator of nematodes, protists, and other small invertebrates. Previous studies have reported that bacterivorous and predatory nematodes are estimated to contribute (directly and indirectly) about 8\% to 19\% of nitrogen mineralization respectively~\cite{bearefungal, neher2001role}. However, \textit{Paratylenchus\_projectus} is a plant pathogenic nematode~\cite{jenkins1956paratylenchus, mcgawley1983reproduction} which primarily parasitizes plant roots. It may lead to stunted plant growth, and potentially reduce plant nitrogen uptake~\cite{claerbout2021thorough} which might be why gFlora and EQO both found it related to PMN. The mechanism of \textit{Paratylenchus\_projectus} affecting PMN needs further study. Moreover, when considering improving one soil function, it is also necessary to consider the impact on other functions (e.g., supporting plant growth), for a holistic understanding of soil health.

We also note that the correlation between the co-response effect of these 5 nematodes and the functional variable PMN is smaller ($r = 0.38$, $p < 0.01$) than that of those bacteria. This suggests that while nematodes do contribute to the production of bioavailable nitrogen in marine clay, their impact may be comparatively modest compared to bacteria. Bacteria, known for their rapid metabolic activities, are often the primary drivers of N-cycling~\cite{zheng2019soil}. They decompose organic matter, releasing nitrogenous compounds that can be further transformed into ammonium through microbial enzymatic actions. This ammonification process is fundamental to the release of nitrogen from organic sources, thereby enriching the sediment with bioavailable nitrogen. While nematodes do contribute to this process through their feeding activities and the processing of organic matter, their impact on N-cycling may be indirect and less pronounced than bacteria. The underlying mechanisms of the interaction between bacteria and nematodes in supporting N-cycling warrant further investigation.

\subsection{The location of important bacteria in the co-occurrence network}

Given the higher significance of bacteria in N-cycling, we then further investigate the location of the 11 most important bacteria in their original co-occurrence network, which shows the entire bacterial community. As can be observed in Fig. \ref{fig:taxa_in_co-occur}, the most important 11 bacteria spread uniformly in the network. Among these important bacteria, 10 are connected to at least one of the other important bacteria. All important bacteria also connect densely with their 54 common neighbors (neighbors shared by at least two important bacteria), which indicates their strong spatial and ecological correlation.

\emph{Clusters} in a network are subsets of nodes that exhibit higher connectivity within themselves than with nodes outside the subset. Nine clusters are detected in this bacterial co-occurrence network using the Louvain algorithm (the network modularity metric, denoted by $Q$, is 0.48)~\cite{blondel2008fast}. We find that these important bacteria do not form an isolated cluster. On the contrary, they are distributed among six clusters, making them better coordinate with other bacteria in the network. Therefore, we hypothesize that the clusters detected simply based on network topology do not necessarily correspond to different functions. In fact, it is likely that multiple functions are performed across different clusters, with various bacteria contributing to different aspects of a specific function. This distribution highlights the complex interplay and functional redundancy within the microbial community, where bacteria from different clusters can collaborate to maintain ecosystem stability and functionality.

\emph{Centrality metrics} are key concepts in network analysis that measure the importance or influence of nodes in a network from different points of view (i.e., definitions of the word importance)~\cite{freeman2002centrality}. Degree centrality measures a node's importance by summing the weights of its direct connections. Closeness centrality measures how quickly a node can reach all other nodes. As shown in Fig. \ref{fig:taxa_in_co-occur}B and C, these important bacteria rank differently in both degree centrality and closeness centrality, which indicates that they play different roles in the network. Some bacteria may be pivotal for facilitating direct interactions with other bacteria (high degree centrality), while others may play a critical role in efficiently transmitting information or resources across the network (high closeness centrality). We found that \textit{Flavisolibacter} (one bacteria that EQO missed) has both high degree and closeness centrality, which suggests it is a central hub in the microbial community, potentially influencing various ecological processes.\textit{ Pajaroellobacter} (another bacteria that EQO missed) also has a relatively high closeness centrality though its degree centrality is low, which implies that it may not have as many direct interactions as \textit{Flavisolibacter}, but it excels in efficiently disseminating information or resources to distant parts of the network. Despite its lower connectivity, the strategic positioning of \textit{Pajaroellobacter} allows it to play a critical role in coordinating microbial activities and maintaining network resilience. Graph convolution thus provides a comprehensive consideration of the roles and importance of all bacteria in discovering the functional co-response group, making it a valuable method for uncovering the intricate dynamics of microbial communities.

%%%%%%%%%%%%%%%%%%%%%%%%%%%%%%%%%%%%%%%%%%%%%%%%%%%%%%%%%%%%%%%%%%%%%%%%%%%%%%%%%%%%%%%%%%%%%%%

\section{Conclusion}
In this paper, we propose gFlora: a new method to discover functional co-response groups from observational data of the soil microbial community. We evaluate gFlora on real-world datasets of soil microbiomes with cross-validation, and results show that our method significantly outperforms the state-of-the-art method. Our main methodological novelty is two-fold: a step of graph convolution, which allows us to learn from the topology of co-occurrence network among microbes, and a new optional regularization step in gFlora which can provide better performance and quicker convergence. 

We show new practical insights on some under-studied taxa related to N-cycling. This is new, non-traditional knowledge, at the level of groups of microorganisms instead of individuals, which is realistic given that taxa function as part of a community. The main limitation is that our functional co-response groups are discovered based on correlations in the data, so causality remains to be validated by experimentation. However, even correlational evidence is important: since validated functional information per taxon is rarely available in the literature and the soil ecological community is complex, this evidence can serve as focused functional hypotheses to test experimentally. We also show new insight on the topological location of functionally important bacteria: they are distributed in the co-occurrence network, hinting at them collaborating across clusters of taxa. However, the $Q$ value of clusters is not very high and the detection of clusters does not mean their formation is statistically separated. Therefore, this is merely an exploratory work to raise hypotheses, which need to be further validated in practice.

gFlora can be easily applied to any other measured soil functions to discover functional groups. In the future, we anticipate developing a comprehensive soil ecological model that incorporates as many soil functions as possible. With this model, we could further investigate the causal relationships between soil functions and those discovered functional groups. This model would provide a more holistic understanding of soil health, helping to balance the benefits and drawbacks of various soil management practices. Besides, both gFlora and EQO only discover the linear relationship between microbiota and the functional variable; non-linear options can also be considered in the future. Furthermore, though measures have been taken against overfitting, the sample size is relatively small thus the risk of overfitting still exists. In future studies, we would apply our model to larger datasets.

%%%%%%%%%%%%%%%%%%%%%%%%%%%%%%%%%%%%%%%%%%%%%%%%%%%%%%%%%%%%%%%%%%%%%%%%%%%%%%%%%%%%%%%%%%%%%%%

\begin{acks}
This work was funded by the Dutch Research Council (NWO, grant number P20-45) in the project SoilProS (Soil biodiversity analysis for sustainability production systems), and by the National Postal Code lottery funded ‘Onder het Maaiveld’ program.
\end{acks}

%%%%%%%%%%%%%%%%%%%%%%%%%%%%%%%%%%%%%%%%%%%%%%%%%%%%%%%%%%%%%%%%%%%%%%%%%%%%%%%%%%%%%%%%%%%%%%%

% AUTHORS' CONTRIBUTIONS:
% Nan: all aspects of the work, incl. research design, implementation, all data visualisation, and writing the manuscript (all sections)
% Merlijn: researched functional information (domain knowledge) of taxa for the discussion of (Important taxa subsection)
% Doina: 2 figures to explain the method, some research design, writing the manuscript (all sections), feedback

%%%%%%%%%%%%%%%%%%%%%%%%%%%%%%%%%%%%%%%%%%%%%%%%%%%%%%%%%%%%%%%%%%%%%%%%%%%%%%%%%%%%%%%%%%%%%%%

\bibliographystyle{ACM-Reference-Format}
\bibliography{ref}

\end{document}